\documentclass[12pt]{article}

\usepackage{multirow}
\usepackage{array}
\usepackage{longtable}

\usepackage{caption}
\usepackage{newtxtext,newtxmath}
\usepackage{float}
\usepackage{dblfloatfix}
\usepackage{graphicx}
\usepackage[letterpaper,margin=1in]{geometry}
\renewenvironment{abstract}
	{\quotation}
	{\endquotation}

\date{}

\makeatletter
\renewcommand{\fnum@figure}{\textbf{Figure \thefigure}}
\renewcommand{\fnum@table}{\textbf{Table \thetable}}
\makeatother

\usepackage{scicite}

\usepackage{url}
\usepackage{caption}

\def\scititle{Context-aware Multimodal AI Reveals Hidden Pathways in Five Centuries of Art Evolution}
\title{\bfseries \boldmath \scititle}

\author{
	Jin~Kim$^{1,2}$,
        Byunghwee~Lee$^{3}$,
        Taekho~You$^{4,5\ast}$,
        Jinhyuk~Yun$^{5,6\ast}$\and
	\small$^{1}$Department of Intelligent Semiconductors, Soongsil University, Seoul, 06978, Republic of Korea.\and
        \small$^{2}$School of Digital Humanities \& Computational Social Sciences, \and \small Korea Advanced Institute of Science and Technology, Daejeon 34141, Republic of Korea.\and
        \small$^{3}$Luddy School of Informatics, Computing, and Engineering, Indiana University, \and \small Bloomington IN 47408, United States of America.\and
        \small$^{4}$Institute for Social Data Science, Pohang University of Science and Technology, \and \small Pohang 37673, Republic of Korea.\and
        \small$^{5}$Center for Digital Humanities \& Computational Social Sciences, \and \small Korea Advanced Institute of Science and Technology, Daejeon 34141, Republic of Korea.\and
        \small$^{6}$School of AI Convergence, Soongsil University, Seoul, 06978, Republic of Korea.\and
	\small$^\ast$Corresponding authors. Email: taekho.you@kaist.ac.kr, jinhyuk.yun@ssu.ac.kr\and
}

\begin{document} 

\maketitle

\begin{abstract} \bfseries \boldmath
The rise of multimodal generative AI is transforming the intersection of technology and art, offering deeper insights into large-scale artwork. Although its creative capabilities have been widely explored, its potential to represent artwork in latent spaces remains underexamined. We use cutting-edge generative AI, specifically Stable Diffusion, to analyze 500 years of Western paintings by extracting two types of latent information with the model: formal aspects (e.g., colors) and contextual aspects (e.g., subject). Our findings reveal that contextual information differentiates between artistic periods, styles, and individual artists more successfully than formal elements. Additionally, using contextual keywords extracted from paintings, we show how artistic expression evolves alongside societal changes. Our generative experiment, infusing prospective contexts into historical artworks, successfully reproduces the evolutionary trajectory of artworks, highlighting the significance of mutual interaction between society and art. This study demonstrates how multimodal AI expands traditional formal analysis by integrating temporal, cultural, and historical contexts.
\end{abstract}

\noindent Understanding what artists depict in their paintings—and how they visually articulate these elements—has long been a central topic in art history. Over the centuries, this topic has been approached primarily through qualitative methods, with art historians carefully analyzing individual works and contextualizing them within broader cultural and historical narratives. However, recent technological advances have fundamentally reshaped how scholars can address such inquiries~\cite{taylor1999fractal, lyu2004digital, hughes2010quantification, kim2014large, elgammal2018shape, sigaki2018history, lee2020dissecting, liu2021understanding, cetinic2022understanding, albright2023paintings, karjus2023compression, yazdani2017quantifying, zhang2022quantification}. The unprecedented availability of large-scale and high-quality digitized images of historical artworks, combined with robust analytical frameworks drawn from statistical physics~\cite{taylor1999fractal, kim2014large, lee2018heterogeneity, koch20101}, information theory~\cite{rigau2007conceptualizing, rigau2008informational, lee2020dissecting, sigaki2018history, karjus2023compression, kim2024diversity}, network science~\cite{schich2014network, fraiberger2018quantifying, elgammal2015quantifying, perc2020beauty}, and machine learning~\cite{gatys2016image, elgammal2018shape, liu2021understanding, lee2024social, stork2024computer}, has broadened the field of quantitative art research. These computational methods complement traditional analyses by offering a systematic means of rigorously testing historical hypotheses, validating or refuting prevailing insights and even uncovering previously unrecognized patterns within vast collections of visual data.

In this regard, substantial progress, including in our previous studies~\cite{lee2020dissecting, lee2018heterogeneity}, has already been made in quantifying what is often referred to as the ``formal elements'' of art---visual features such as line, shape, form, color, texture, space, and contrast---and in analyzing compositional strategies that govern how these elements are organized within artworks~\cite{feldman1971varieties, rose2019art, arnheim1954art}. Computational studies have provided a deeper understanding of aesthetic styles and trends, revealing structural patterns that extend across periods and regions. These studies have also elucidated how formal elements contribute to the perception of balance, harmony, and visual interest. As an illustrative example, early studies focused on the fundamental geometric properties of abstract paintings, \textit{e.g.}, Pollock's drip paintings showing fractal patterns~\cite{taylor1999fractal}. Through further development of fractionality analysis, information-theoretic approaches have found the historical pathway of formal elements by measuring the complexity and entropy of artworks~\cite{sigaki2018history}. Another study analyzed the temporal evolution of color contrast in paintings, showing increasing diversity across centuries and how formal elements continue to shape artistic expression~\cite{lee2018heterogeneity}. Studies have also examined the partition patterns in landscape paintings from a compositional perspective~\cite{lee2020dissecting}. Modern convolutional neural networks have also been used to extract visual features from paintings, revealing distinct clustering patterns among artists and styles in the principal component space~\cite{elgammal2018shape}. However, while these advances have refined our understanding of form and structure, they represent only one dimension of what art conveys: formal elements.

An equally critical but relatively underexplored aspect of art in quantitative art research is the \emph{contextual} aspect. Contextual elements relate to the subjects, objects, themes, and narratives depicted in artworks, offering insights into the cultural, historical, and intellectual environments in which they were produced~\cite{albright2023paintings}. These elements capture the essence of what artists choose to represent---whether symbolic, religious, or everyday scenes---and how these choices reflect the evolving concerns and priorities of their time. Despite their importance, the macroscopic study of contextual elements has been limited by the challenges of extracting contextual information from visual data, where variability in style, medium, and symbolic representation complicates large-scale analysis.

Recent developments in multimodal artificial intelligence (AI) offer a promising pathway for overcoming these challenges~\cite{caldas2020generative, zhan2023multimodal,zhao2024m}. By integrating information from multiple modalities---such as images and textual descriptions---multimodal AI models can learn latent vector representations that encode not only the formal properties of an artwork but also its contextual elements. These models provide a unified framework for analyzing art at both the structural and thematic levels by bridging the gap between visual representation and semantic and contextual information.

In this study, we explored the representation of visual art within multimodal AI's latent spaces, with a particular focus on their ability to capture the temporal evolution of art. Specifically, we examined whether these latent spaces, enriched with multimodal information, can capture chronological distinctions with greater resolution than those derived solely from formal elements. We pose the following question: Can contextual embeddings informed by multimodal AI distinguish artworks across historical periods and stylistic movements more effectively? To address this, we propose a comparative framework that evaluates the relative efficacy of formal and contextual latent spaces in characterizing temporal trends in visual art. Through this analysis, we sought to demonstrate the unique potential of multimodal AI in revealing previously inaccessible dimensions of artistic change, highlighting the value of content-driven approaches in advancing our understanding of art history's temporal dynamics and cultural significance.

Specifically, we used the Stable Diffusion Model (SDM) to analyze 500 years of Western paintings~\cite{rombach2022high, mao2017deepart}, where images are mapped onto latent vectors through its built-in autoencoder and CLIP, along with 72,447 paintings consisting of 2,354 painters and 128 conventional style periods. The autoencoder compresses high-dimensional images onto a low-dimensional latent space while preserving essential visual information, while CLIP simultaneously encodes visual and contextual features. We generated two distinct representation vectors for each painting: \textit{A-vectors} from the autoencoder and \textit{C-vectors} from CLIP. Our comparative analysis between two vectors highlights the importance of contextual elements captured by C-vectors, demonstrating their superior discriminative power among periods, artistic styles, and artists compared to A-vectors.

We then extract seminal keywords from paintings using modern language models. Using these contextual keywords, we traced changes in the motifs and subjects of Western art history and found that artistic progress is closely associated with societal and technical transitions. We also attempted to reconstruct the evolution of art history through temporal diffusion experiments, in which we infused contextual elements extracted from subsequent centuries into existing paintings. Our findings provide quantitative evidence that broader societal changes, reflected in contextual features, are key drivers in the emergence of new artistic expressions, styles, and themes. This approach offers a novel method for analyzing paintings by integrating formal and contextual dimensions and demonstrates the value of AI in uncovering deeper contextual insights beyond traditional image-based analyses.

\begin{figure}[H]
    \centering
    \includegraphics[width = 0.65\textwidth]{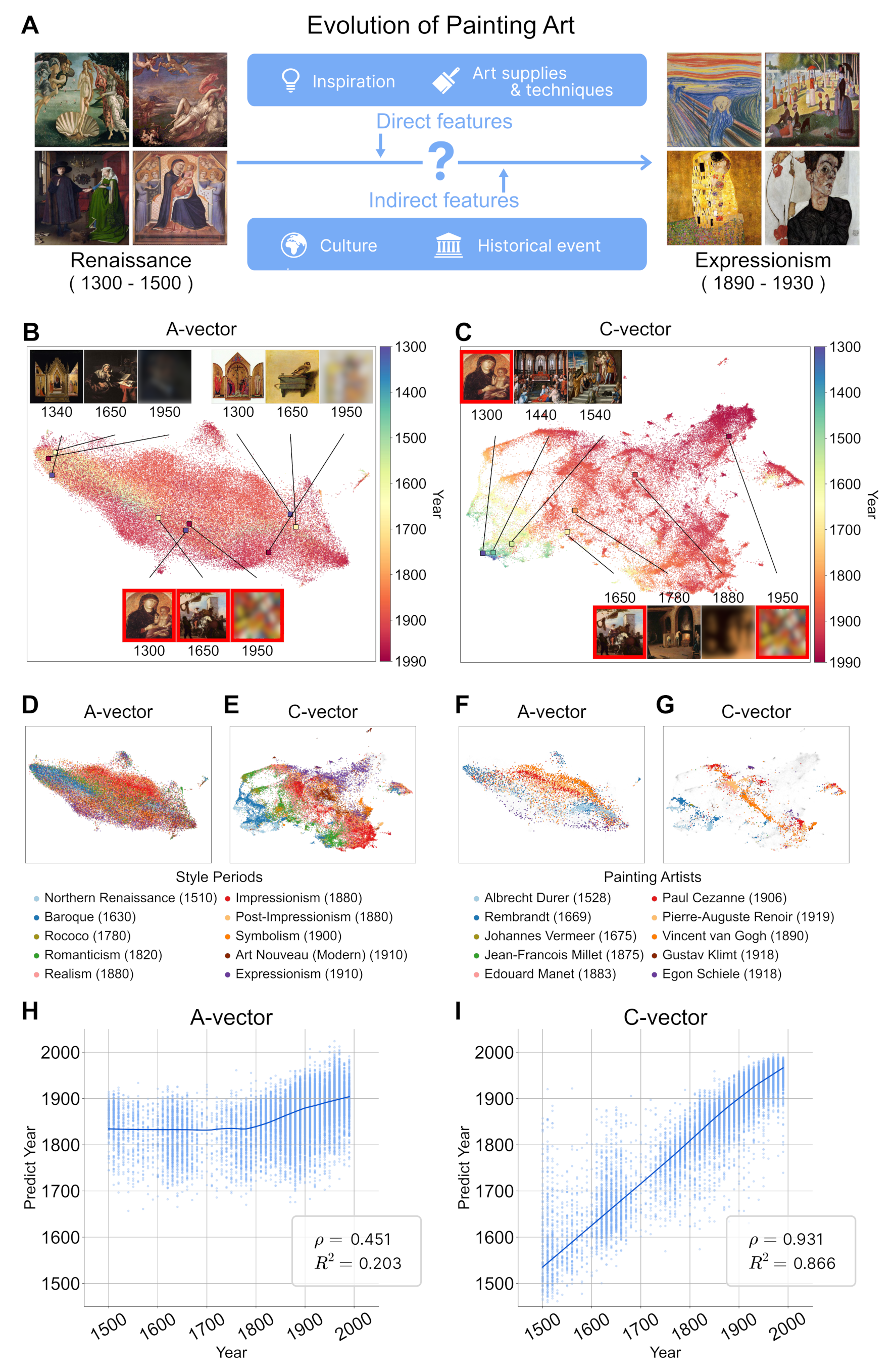}
    \caption[]{\textbf{Understanding the Evolution of Western Paintings with Image Embedding} (A) The evolution of paintings has paralleled human evolution through historical events, technological advancements, and cultural developments, yet contextual information is commonly disregarded in data-scientific approaches to art history. (B--G) A two-dimensional (2D) projection of 72,447 Western paintings was obtained using Uniform Manifold Approximation and Projection (UMAP) with the encoded vectors~\cite{mcinnes2018umap}, where each dot represents a painting. To emphasize the importance of contextual content in Western paintings, we encoded the paintings using an (B, D, F) autoencoder (A-vector) and (C, E, G) CLIP (C-vector)~\cite{kingma2019introduction,radford2021learning}. Note that dot colors indicate painting years. A-vectors show mixed distribution, making it difficult to distinguish painting years (B), while C-vectors effectively differentiate them (C). For instance, red-bordered paintings sampled from different periods are clustered in the center of the A-vector space but are well-separated in the C-vector space. This pattern holds for (D and E) the 10 most frequent style periods and (F and G) 10 seminal artists (H and I). The year displayed next to the artist name is their death year, whereas those next to the style period is the most frequent painting year. To test the expressibility of each vector, we train regression models using XGBoost~\cite{chen2016xgboost} to predict painting years from embedded vectors (see Materials and Methods). Here, solid lines are drawn with locally weighted regression~\cite{cleveland1979robust}. A-vectors demonstrate limited predictability ($R^2=0.203$, Pearson $\rho=0.451$), while C-vectors show remarkably high predictability ($R^2=0.866$, Pearson $\rho=0.931$).}  
    \label{fig:fig1}
\end{figure} 

\subsection*{Time predictability of latent vectors}\label{subsec:predictability}

Western painting is conventionally categorized by artistic movements with distinctive characteristics. To fully comprehend these characteristics, we must understand not only the formal elements but also the internal and external contextual dimensions. The evolution of artistic movements can be conceptualized as a reflection of each era's cultural, social, and philosophical contexts (Fig.~\ref{fig:fig1}A). For instance, the development of new painting tools and steam-powered transportation facilitated the emergence of landscape paintings, whereas the Industrial Revolution influenced the rise of impressionistic works~\cite{albright2023paintings}. Building on this foundation, we examined how to capture the evolution of paintings by extracting both formal and contextual information.

One advantage of encoders in AI models is that they learn important features while reducing the dimension of the original inputs. We employed an autoencoder and CLIP within the SDM to embed painting images into two distinct latent spaces, producing A- and C-vectors that encode formal and contextual features, respectively (see Materials and Methods). The Uniform Manifold Approximation and Projection (UMAP)---a dimensionality reduction method that groups neighboring paintings within the manifold structure of a high-dimensional space~\cite{mcinnes2018umap}---revealed that A- and C-vectors form distinct spatial distributions. The UMAP of the A-vectors exhibited a clustered pattern based on visual characteristics but did not show distinct clusters by painting year (Figs.~\ref{fig:fig1}B and S5). Instead, paintings from the 19th and 20th centuries were widely distributed, suggesting that the A-vectors did not effectively capture the temporal characteristics of paintings. In contrast, the UMAP of the C-vectors displayed a progressive trend with time (Figs. ~\ref{fig:fig1}C and S5), from 14th-century paintings occupying the bottom left to 20th-century paintings positioned in the upper right. This continuous progression from the bottom left to upper right reveals a chronological pattern across centuries. Notably, paintings from the 19th and 20th centuries were widely distributed in the C-vectors (Fig.~\ref{fig:fig1}C), but they remained distinct without mixing with other periods.

One may argue that this can be attributed to the larger number of paintings in recent periods (Fig.~S1), yet we consistently observed limited expressibility of the A-vectors. For example, although we can observe some clustered structures in the A-vectors when grouping paintings by style periods and artists, the degree of clustering in these groups is considerably weaker than that in C-vectors (Figs.~\ref{fig:fig1}D, F, S6, and S7). Despite the significant divergence in the number of paintings across different artists and style periods (Figs.~S2~and~S3), the UMAP of the C-vectors also demonstrated well-bounded clustering according to style periods and painting artists (Figs.~\ref{fig:fig1}E, G, S6, and S7). This is further confirmed by the distances between paintings in the A- and C-vectors (Fig.~S11). C-vectors showed a significantly larger separation between different authors (mean distance: $1.138$ versus $0.959$ for different and same authors, respectively, with $18.7\%$ difference) than A-vectors (mean distance: $153.010$ versus $143.753$ for different and same authors, respectively, with $6.4\%$ difference). This better discriminative power was also confirmed by the Kolmogorov-Smirnov (KS) test, where C-vectors showed a substantially higher KS statistic between the same and different author distributions ($0.614$) compared to A-vectors ($0.206$). C-vectors also exhibited more consistent measurements across different styles (KS statistic: 0.330 versus 0.053 for C- and A-vectors, respectively; mean distance: 1.122 versus 1.039 for C-vectors; 149.150 versus 146.631 for A-vectors).

We also observed that related artists and styles clustered close to each other. For example, the paintings of \textit{Gustav Klimt} and \textit{Egon Schiele} are closely positioned in the C-vector space, reflecting their historical and stylistic connections (Figs.~\ref{fig:fig1}G and S7). The two artists lived in Vienna in the early 20th century and profoundly influenced each other's works. However, the paintings of \textit{Pierre-Auguste Renoir} are separate from these two artists' works in Fig.~\ref{fig:fig1}G, reflecting the French impressionist's distinct artistic approach and geographical separation from the Viennese artists. This pattern was not observed in A-vectors, where Klimt and Renoir were located together, whereas Schiele and Klimt were separated from each other (Figs.~\ref{fig:fig1}F and S7). Note that we obtained similar patterns with another dimensionality reduction method (t-SNE; see Figs.~S8, S9, and~S10), demonstrating the robustness of our findings.

To quantify the temporal resolution of the A- and C-vectors, we evaluated their capacity to predict the historical period of artworks. We employed the XGBoost regression model, which is widely used in machine learning tasks~\cite{chen2016xgboost}. For each vector type, we trained separate models using 70\% of the original paintings as training data, with their respective latent vectors as input features to predict the year of creation. We then assessed the predictability of the models using the remaining 30\% of paintings (see Materials and Methods for further details). The locally weighted regression line~\cite{cleveland1979robust} suggests that the A-vector model predominantly misclassified early paintings (from the 16th to 18th centuries) as 19th-century works ($R^2 = 0.207$; Fig.~\ref{fig:fig1}H). The model also predicted numerous 20th-century paintings as 19th-century works. Meanwhile, the C-vector model consistently showed more accurate creation year predictions ($R^2=0.869$; Fig.~\ref{fig:fig1}I). In short, the C-vectors exhibited superior performance in predicting the creation year of artworks compared to the A-vectors.

\subsection*{Unveiling latent information encoded in vectors}

In the previous section, we observed that C-vectors exhibit greater predictive power than A-vectors. Thus, a natural question arises: what makes C-vectors more expressible than A-vectors? To answer this question, we conducted an in-depth analysis of the encoded information, particularly for paintings, within the A- and C-vectors to unveil the factors that make C-vectors more interpretable. Because both the autoencoder and CLIP models were trained using general image datasets, we first performed principal component analysis (PCA) on the A- and C-vectors to extract the key features of paintings. We then investigated how paintings vary in style and context along each PC axis to interpret what each axis encodes.

\begin{figure}[H]
    \centering
    \includegraphics[width = 0.7\textwidth]{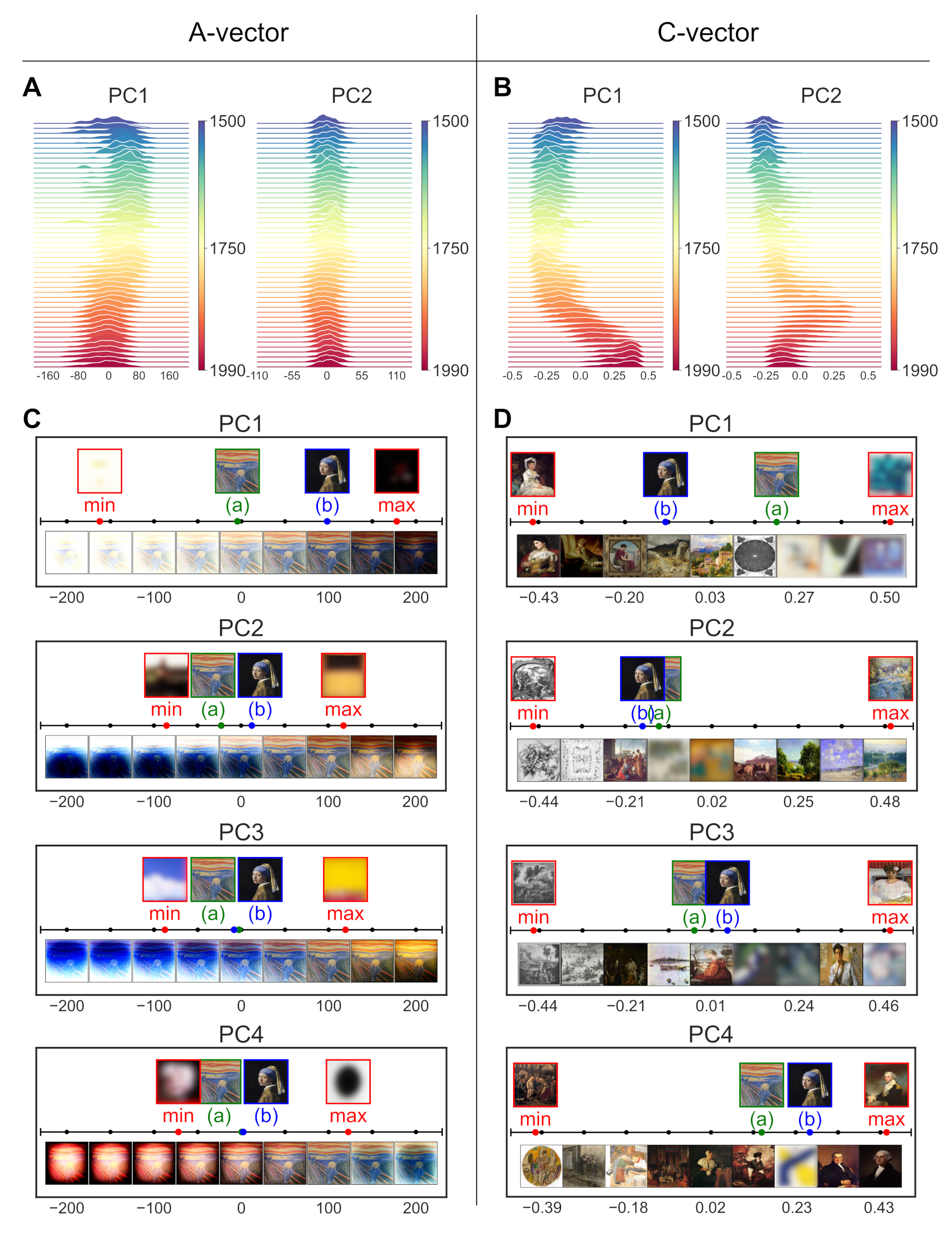}
    \caption{\textbf{Principal Component Analysis (PCA) Reveals Latent Information Encoded in Embedded Vectors.} To explore the expressibility  gap between A- and C-vectors, we conducted PCA to extract principal information from each vector representation. The first two components of each vector space revealed that (A) A-vectors show minimal variation, while (B) C-vectors demonstrate substantial temporal differentiation. For (A) and (B), the x-axis is bounded by the maximum and minimum projected values of paintings on the PC. (C) We obtained modified A-vectors of Munch's ``The Scream'' using vector analogy: $ \textbf{v}_{\mbox{new}} = \textbf{v}_{\mbox{original}} + d \cdot \textbf{PC}_{\textbf{i}}$, where $\textbf{v}_{\mbox{original}}$ represents the embedded vector of ``The Scream'' and $\textbf{PC}_{\textbf{i}}$ denotes the normalized i-th PC vector. Images were then generated using the SDM's autoencoder with these modified vectors~\cite{rombach2022high}. The resulting images show that the first four PCs of the A-vector primarily represent visual composition elements: 1) brightness, 2) vertical brightness composition, 3) hue (blue to yellow), and 4) highlight distribution. The original vector of ``The Scream'' is placed near the zero point, while Vermeer's ``Girl with a Pearl Earring'' has a high PC1 value. (D) For comparison, we retrieved paintings based on their C-vector's projected values and distances on each PC (see Materials and Methods for detailed selection criteria), as CLIP's encoder-only architecture lacks image generation ability. We observed that images mainly vary in context along the PCs, as exemplified by transitioning from portraits to abstract compositions in PC1.} 
    \label{fig:fig2}
\end{figure}

We first projected paintings onto the principal components (PCs) by computing the dot product between the latent vectors of paintings and each PC. The distributions of the projected values on the first- and second-largest PCs (PC1 and PC2, respectively) of the A-vectors initially appeared to show little evidence of temporal patterns (Fig.~\ref{fig:fig2}A). We noticed subtle temporal variations, with a modest increase during 1500--1550, followed by a plateau period between 1550 and 1750, before a gradual decline throughout 1750--1990 for PC1; meanwhile, the PC2 distribution remained temporally consistent. We observed a similar temporal stability across the remaining PC axes of the A-vector (Fig.~S13). It should be noted that the eigenvalues (i.e., explained variance ratios) of the PCs for both the A- and C-vectors exhibited steeply declining distributions (Fig.~S12); the largest PCs (corresponding to the largest eigenvalues) largely explain the vital information that the vectors encode for paintings. As PC1 has a high explained variance ratio (0.190), which is almost six times that of PC2 (0.034), PC1 shows the significant expressibility of paintings for the A-vector. Therefore, although the PCs of the A-vector effectively characterize certain aspects of paintings, they lack a clear temporal correspondence.

In contrast, the C-vector demonstrated more drastic temporal variations in PCs over time (Fig.~\ref{fig:fig2}B). On the PC1 axis of the C-vector, the projected values showed a rapid increase from the late 1800s and then saturated from 1970. PC2 peaked in the late 1800s and then declined, displaying an inverse temporal trajectory compared to PC1. Similar distinct temporal patterns were observed across the remaining PC axes of the C-vector (Fig.~S13). Note that these two PCs maintained relatively low explained variance ratios (0.063 and 0.050), and the explained variance in the C-vectors decreased more gradually than that of A-vectors (Fig.~S12), suggesting that the information in the C-vectors was more evenly distributed across the principal dimensions of the artwork. In short, these temporal patterns indicated that the features captured by the C-vectors exhibited stronger temporal sensitivity (Fig.~S13).

Next, we investigated the aspects of paintings encoded by each PC by modifying the PC values of paintings for both vector types. The autoencoder comprised both encoder and decoder models, enabling us to generate images directly from the modified A-vectors. In contrast, CLIP is an encoder-only model, which prevented the generation of images directly from the C-vectors. Consequently, we employed different analytical strategies for each vector type. For the A-vectors, we generated new latent vectors by manipulating the paintings along the PC directions as follows:

\begin{equation}
    \textbf{v}_{\mbox{new}} = \textbf{v}_{\mbox{original}} + d \cdot \textbf{PC}_{\textbf{i}},
    \label{equ:equ1}
\end{equation}

\noindent where $\textbf{v}_{\mbox{original}}$ is the A-vector, $\textbf{PC}_\textbf{i}$ is the \textbf{i}-th principal component, and $d$ is a scalar parameter that adjusts the position along the PC axis ranging in (-200, 200). We then generated images of $\textbf{v}_{\mbox{new}}$ using the decoder part of the autoencoder. This method allows us to monitor how a painting changes as we adjust its position along the PC axis, revealing which aspects of the artwork each principal component represents. As an illustrative example, we show modified versions of Edward Munch's ``The Scream'' by adjusting $\textbf{v}_{\mbox{original}}$ in Fig. ~\ref{fig:fig2}C. For the C-vector, because there is no decoding method for $\textbf{v}_{\mbox{new}}$, we found the painting closest to each point along the PC axis (see Materials and Methods for details).

Then, what does each PC axis represent? We found that each PC of the A-vectors captures specific visual elements, such as brightness, hue, and composition (Fig.~\ref{fig:fig2}C). The PC1 axis predominantly encodes the overall brightness, where paintings with brighter tones have a negative value and darker compositions have a positive value. These brightness variations manifest more intensely at the peripheral boundaries of the paintings than in their central regions, corresponding to the common artistic practice in which foreground subjects are spotlighted more than background elements (see PC4). As an illustrative example, \textit{Johannes Vermeer}'s ``Girl with a Pearl Earring,'' which features the main subject (girl) centered against a dark background, is positioned at ${\sim}98.27$ on the PC1 axis, whereas \textit{Edvard Munch}'s ``The Scream'' occupies a more neutral value (${\sim}{-}4.53$). PC2 captured the horizontal compositional elements of the paintings. The minimal abstract work ``No.9 (Dark over Light Earth)'' by \textit{Mark Rothko}, featuring black blocks positioned upon yellow, is located at the maximum value (${\sim}117.45$) of the PC2 axis, whereas at the opposite extreme (minimum, ${\sim}{-}85.48$), ``Portrait Of Orca Bates 1989'' by \textit{Jamie Wyeth}, displays a composition with bright areas contrasted by darker elements at the bottom. PC3 represent colors---blue to orange. PC4 also emphasize different compositions and highlight patterns. In summary, the A-vector mainly encodes the ``formal element'' of artworks, which is conventionally investigated by data-driven artwork studies~\cite{feldman1971varieties, rose2019art, arnheim1954art, lee2020dissecting, lee2018heterogeneity}.

In contrast to the A-vector, the PC axes of the C-vector primarily captured the semantic and contextual elements of paintings, \textit{e.g.}, subject matter, historical context, and stylistic movements (Fig.~\ref{fig:fig2}D). For instance, on the PC1 axis, paintings featuring human subjects and portraits occupied the leftmost negative values (minimum PC1), whereas more geometric or abstract compositions had the rightmost positive values (maximum PC1). This result is consistent with our observations of the PC1 projected values in Fig.~\ref{fig:fig2}B, showing a rapid increase from the late 1800s to the late 1900s. PC1 thus reflects the historical shift towards abstract art during this period. The PC2 axis distinguishes between different subject domains, separating human figures (negative values) from landscape paintings (positive values). For example, both ``The Scream'' and ``Girl with a Pearl Earring'' occupy negative values on the PC2 axis. As we observed increasing PC2 values from the late 1700s, with a peak in the late 1800s (see Fig.~\ref{fig:fig2}B), this pattern aligns with the historical proliferation of landscape painting, driven by technological advances, \textit{e.g.}, steam locomotives and tube pigments~\cite{nga_impressionist, sothebys_impressionism}. PC3 introduces further nuance, differentiating between simple portraits and more complex compositional scenes in which human subjects appear within elaborate backgrounds. These distinctions suggest that C-vectors encode not only visual objects but also deeper contextual relationships within artistic traditions. Some PCs capture more complex variations that humans cannot easily capture; \textit{e.g.}, PC4 does not follow a simple regularity but reflects a more intricate structure, capturing hidden context gradually transitioned by time that even humans hardly perceive (Fig.~S13). This demonstrates the further potential of the AI model in art history studies. Despite this complexity, images with smaller PC4 values generally display more realistic features than those with larger PC4 values, indicating that this dimension may capture the degree of artistic realism versus stylization.

These findings reveal the differences in the information embedded in the A- and C-vectors and demonstrate that each vector characterizes distinct temporal changes in the features of artworks. The A-vector mainly captures features of the structure of the paintings, \textit{i.e.}, formal elements, whereas the C-vector primarily encodes information about the objects, \textit{i.e.}, contextual elements. As shown in Fig.~\ref{fig:fig1}, the A-vector has limited expressibility for the creation year, artist, and style period of paintings. Therefore, we may conclude that the formal elements--- thepredominantly used features for the data-driven artwork research---are insufficient. Instead, C-vectors, which effectively capture more contextual information, allow us to detect distinctions in the creation year, artist, and style of paintings. In other words, incorporating contextual information with the social and historical backgrounds of paintings is essential for a deeper understanding of the evolution of art.

\subsection*{Contextual evolution of paintings}\label{subsec:contextual_information}

In the previous section, we demonstrated that C-vectors effectively capture contextual information in paintings and better reflect temporal changes in artworks. This finding naturally leads us to question whether the content and contextual information depicted in paintings genuinely reflect societal changes. The literature supports the idea that social change can lead to the emergence of new artistic styles. For example, scholars have identified how air pollution in the 19th century contributed to the emergence of impressionism~\cite{albright2023paintings}. Combined with our findings, we assume that extracting human-understandable contexts encoded in C-vectors from paintings should enable us to trace social changes. Therefore, we designed an analysis that extracted contextual keywords from paintings and analyzed their temporal changes.

We assume that a text prompt represents an image most accurately among alternatives if the original image can be regenerated from the prompt in the most similar form. Thus, we inferred the generative prompts by CLIP Interrogator~\cite{clipInterrogatorGthub}, combining the CLIP~\cite{kingma2019introduction} and BLIP~\cite{li2023blip} models, in which the prompt reproduces the original image as much as the SDM can (see Materials and Methods). We then separated each prompt into (1-gram) keywords using delimiters (e.g., spaces and commas) and calculated their normalized frequencies, where the frequency of each word was divided by the total sum of all word frequencies within its respective decade.

\begin{figure}[H]
    \centering
    \includegraphics[width=0.7\textwidth]{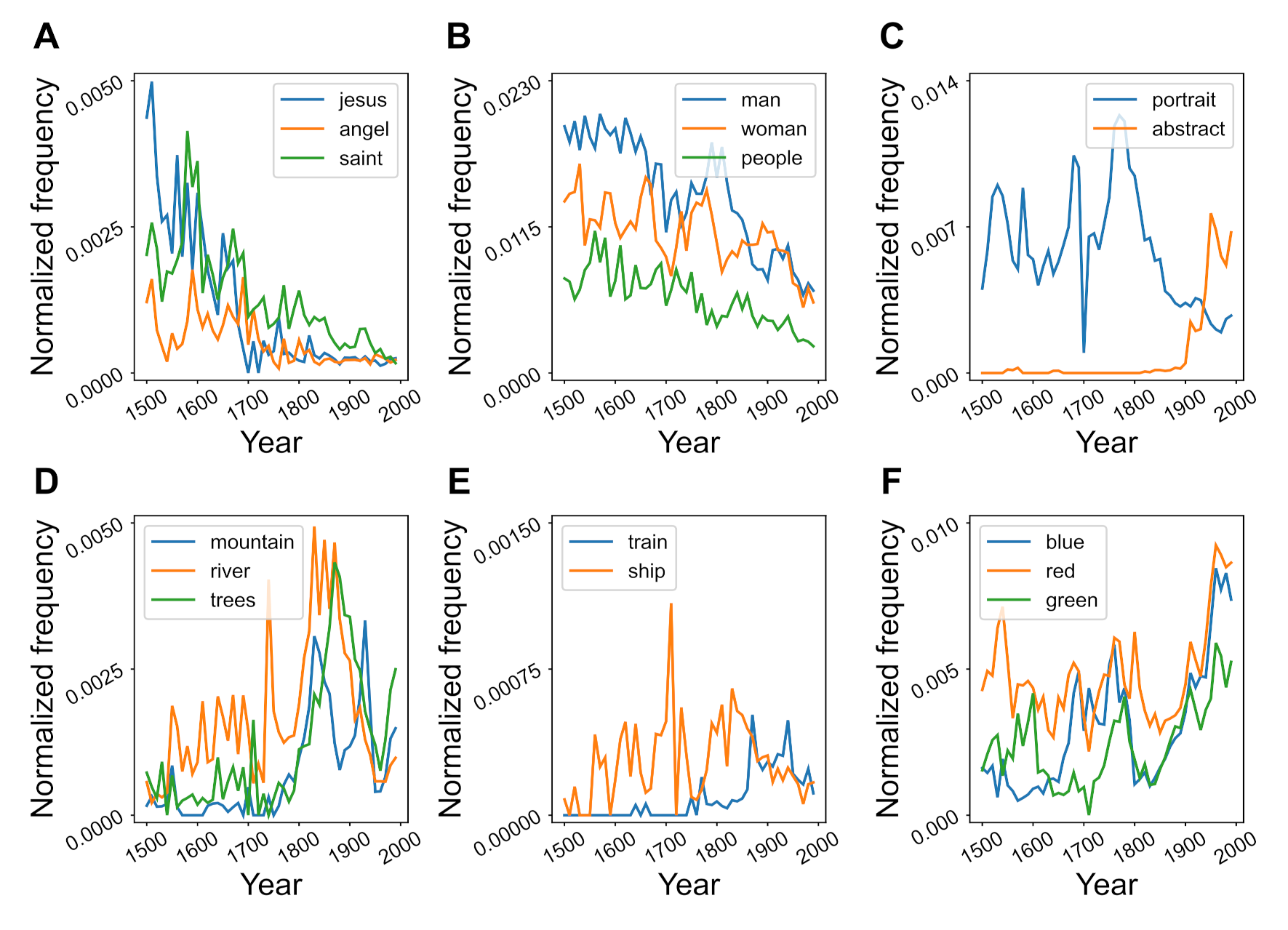}
        \caption{\textbf{Generative Prompt Outlines Temporal Evolution of Western Art}. (A) Religious words show decreasing trends over time. (B) Similarly, human subject descriptors also gradually decreased. (C) Comparative analysis of artistic style descriptors \textit{abstract} and \textit{portrait}, highlighting the transition in representational approaches. (D) Natural features abruptly increased around the 1800s, likely reflecting the development of tube colors and steam locomotive trains enhancing the mobility of painters. (E) We also observed notable changes in transportation-related keywords. For example, the term \textit{train} steeply increased following the invention of steam locomotives in the 1800s. (F) The rise of simple color words indicates an evolution towards an abstract style in the early 1900s. These findings highlight how multimodal AI models can effectively illustrate shifts in artistic expression and content in response to technological, societal, and stylistic changes throughout history.}
    \label{fig:fig3}
\end{figure}

Changes in representative keywords over time revealed intriguing patterns in the painting content. We first detected a substantial decline in religious keywords, such as \textit{jesus}, \textit{angel}, and \textit{saint}, which became rare after 1700 (Fig.~\ref{fig:fig3}A). This pattern reflects society's gradual disengagement from religion over time. Meanwhile, human-related keywords (\textit{man}, \textit{woman}, and \textit{people}) also decreased yet maintained a considerable portion (Fig.~\ref{fig:fig3}B). In this regard, our comparative analysis of artistic style descriptors revealed a pronounced increase in the term \textit{abstract}, coupled with a corresponding decline in \textit{portrait}, which primarily depicts humans. This trend emphasizes the fundamental transition in representational practices throughout art history (Fig.~\ref{fig:fig3}C; see Fig.~\ref{fig:fig2}B and D also for the increment of PC1, showing an increasing trend of abstract arts). We observed a significant increase in keywords related to landscape painting, such as \textit{mountain}, \textit{river}, and \textit{trees}, starting from the late 1700s and peaking in the late 1800s (Fig.~\ref{fig:fig3}D); this trend corresponds with technological advances, such as steam locomotives and tube pigments, as reflected in our PCA results (see the steep increase in PC2 in Fig.~\ref{fig:fig2}B and D). We further confirmed this pattern through the increment of the keyword \textit{train} (Fig.~\ref{fig:fig3}E). Primary colors also increased significantly since the 1800s, coinciding with the emergence and rise of abstract styles (Fig.~\ref{fig:fig3}F). We also identified the most increased and decreased keywords over the five centuries through regression analysis (see Tables~S1 and~S2, respectively, along with Fig.~S14). This analysis validates the results shown in Fig. ~\ref{fig:fig3}. For instance, \textit{abstract} ranked as the 3rd most increased keyword, whereas color-related terms ranked highly (e.g., \textit{blue} as 2nd, \textit{white} as 4th, \textit{yellow} as 5th, \textit{color} as 10th, \textit{green} as 12th, and \textit{red} as 17th). On the other side, human-related terms (e.g., \textit{man}, \textit{people}, \textit{woman}, and \textit{portrait}) and religious terms (e.g., \textit{jesus}, \textit{saint}, and \textit{angels}) ranked as the most decreased keywords.

In summary, we observed notable evidence that social changes, including the evolution of painting tools, technological advances, and social events, largely influenced the formation and development of paintings. These external factors shaped artistic expression and thematic choices across different historical periods, demonstrating a close relationship between societal transformation and artistic evolution.

\subsection*{Future contextual information navigates forthcoming paintings}\label{subsec:diffusion_test}

Our previous findings revealed that contextual information extracted from paintings is closely related to societal changes and that this information can successfully regenerate images similar to the original ones (Fig. ~\ref{fig:fig3}~and~S4). The robust expressibility of C-vectors also suggests their potential utility beyond reconstruction. Because contextual features effectively encode temporal progression in artistic development, we hypothesized that these vectors might facilitate the prediction of the evolutionary trajectory of Western paintings. This relationship led us to the ultimate question: can we generate forthcoming paintings with current images infused with future contextual information? Unfortunately, we cannot determine whether the generated artistic style truly represents the future. Instead, we examined historical records to find evidence supporting our hypothesis. We investigated whether images generated from paintings from a specific period, infused with contextual information from subsequent periods, showed forward temporal movement. We determined this by examining whether the generated images were classified as belonging to these subsequent periods using our year prediction model based on the C-vector, as shown in Fig.~\ref{fig:fig1}.

To explore this possibility, we designed the following generative test. First, we selected the 77 representative keywords for each century based on their TF-IDF values, where an artwork corresponds to a document (see Materials and Methods for detailed selection criteria; see Table~S5 also for the full selected keywords list). These keywords were derived from paintings from each century using the same text extraction method described in the previous section. We selected 77 words because this corresponds to the maximum token limit that can be included in the prompt for image generation using our SDM~\cite{radford2021learning}. Because the minimum number of tokens from the 77 words was exactly 77, selecting this number guaranteed at least 77 tokens (see Table~S5 for the full word list). This approach allowed us to maximize the information content within the constraints of the model architecture, thereby ensuring optimal representation of period-specific artistic vocabulary. We then generated space-separated prompts for each century (see Materials and Methods for details; testing of alternative comma-separated prompts is also presented in Fig.~S16). Second, we randomly sampled 500 paintings from each century. Third, for paintings from the \textit{t}-th century, we generated paintings with the keywords of the (\textit{t+1)}-th century using the SDM~\cite{rombach2022high}. For the comparative null model, we also generated painting images from the same $t$-th century images using the same SDM but without providing any keywords. We refer to the former (with keywords) as \textit{future-directed paintings} and the latter (without keywords) as \textit{random diffusion paintings}. The SDM generates an image by guiding the diffusion direction based on the text input. Therefore, we assume that the keywords of the (\textit{t+1})-th century guide the diffusion process towards the future from the original image, where an image without any text input is assumed to be randomly directed. We repeated this process by varying the \texttt{step} parameter to modify the degree of perturbation. This process allowed us to observe both the evolution of paintings and degree of transformation influenced by the text input (see Materials and Methods).

\begin{figure}[H]
    \centering
    \includegraphics[width = 0.7\textwidth]{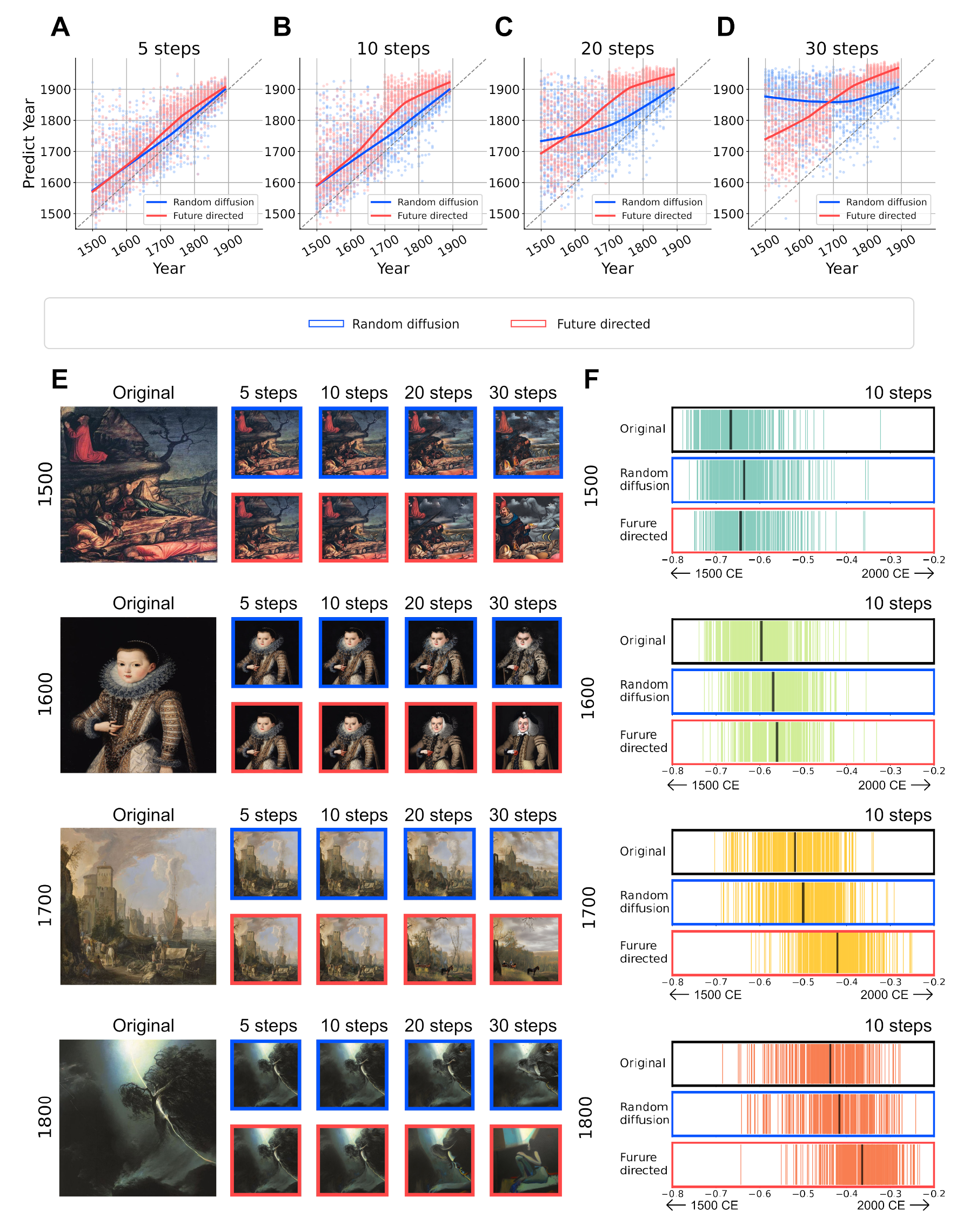}
    \caption{\textbf{Replicating Evolutionary Trajectories of Western Paintings.} To verify our findings on the role of context in understanding the evolution of art, we designed a simple generative experiment with an image-to-image diffusion model using different prompt guidance: one used a null prompt (random-diffusion), while the other used representative keywords of the next century, \textit{e.g.}, keywords of the 1800s for the paintings of the 1700s (future-directed; see Materials and Methods). (A)--(D) Using the generated images, we first estimated the painting year for each using the C-vector regression model shown in Fig.~\ref{fig:fig1}, which showed that the future-directed images (red lines) tend to be predicted consistently $\simeq 100$ years ahead of their source painting, demonstrating a systematic temporal shift in artistic characteristics. In contrast, the random-diffused images (blue lines) showed a distinctive convergence towards ${\sim}1900$ as the diffusion steps increased. Here, solid lines are drawn with locally weighted regression~\cite{cleveland1979robust}, and steps control the relative noise level, where larger steps reduce the consistency with the original image by applying stronger noise (see Materials and Methods). (E) Sampled images for each period, showing that diffusion steps do not alter the large formal elements of paintings for both random-diffused (blue-boxed) and future-directed (red-boxed) images. (F) Spectrum of the paintings. The color represents the original period of the painting, and the position reflects the projected value of C-vector onto the temporal axis $\textbf{v}_{1500s \rightarrow 1900s}$ for original (black-boxed), random-diffused (blue-boxed), and future-directed (red-boxed) images at step 1  (see Materials and Methods). For all panels, we randomly sampled 500 images from each period to reduce bias from data imbalance.}
    \label{fig:fig4}
\end{figure}

The results show that future-directed paintings were more accurately classified as works from the subsequent century relative to their original creation time compared to random diffusion. Notably, as the number of diffusion steps increased, the images generated by the random diffusion model tended to be classified as paintings from the 20th century, where modern abstract art became more prevalent. We attribute this classification tendency to the use of randomness in modern artistic practices, as exemplified by highly experimental artists such as \textit{Jackson Pollock}~\cite{kim2014large}. We can reasonably infer that the model has difficulty differentiating between pure random patterns and intentional artistic randomness. Our investigation confirmed this limitation; when we input pure white noise patterns into the year prediction model, they consistently predicted years around the mid-20th century (Fig.~S15). By contrast, the future-directed model consistently predicted paintings to be from the century immediately following the original artwork, presenting a more plausible temporal progression (Fig~\ref{fig:fig4}A--D).

An interesting observation from the generated images---both future-directed and random-diffusion---is that they closely resemble the original paintings in terms of their formal aspects (Fig.\ref{fig:fig4}E). The paintings retain the same outlines, colors, and compositions, indicating that their \emph{formal} characteristics remain largely unchanged. However, upon closer inspection, subtle differences in content emerged. For instance, a sample portrait painting from the 1600s in Fig.\ref{fig:fig4}E underwent a dramatic transformation in the hairstyle and clothing of the portrait subject as the diffusion steps increased. At 30 steps, the subject appeared to adopt a silver-colored wig, which corresponds to the wig vogue in the 18th century~\cite{kwass2006big}; note that the term `wig' is also the top keyword of the 18th century (Table~S5). Additionally, a sample painting from the 1800s transformed from a tree into an abstract human figure, reminiscent of surrealist art. These findings suggest that \emph{contextual} information, represented by keywords extracted from certain periods, can significantly influence the content of a painting, making it resemble a future artistic style while preserving its formal attributes.

Finally, from an alternative perspective, we confirmed that future-directed paintings exhibited a stronger tendency to move towards the future. We calculated the temporal axis $\textbf{v}_{1500s \rightarrow 1900s}$ as the difference between the average C-vectors of all paintings from the 1500s and those from the 1900s (see Materials and Methods). When projecting paintings onto the vector axis capturing the temporal variation from 1500 to 1900 in the C-vector space, future-directed paintings consistently shifted towards higher average values compared to random-diffusion paintings and original paintings, except for those from 1500, as shown in Fig.~\ref{fig:fig4}F. Random-diffusion paintings also exhibited higher average values than original paintings, although the difference was marginal. Note that the results are robust regardless of the prompt format (see Materials and Methods and Fig.~S16).

In summary, we confirmed that future paintings can be induced through experiments that infuse future context into a painting without altering the formal elements. As discussed in the previous sections, this verifies that context has a significant influence on artistic development. These findings also verify the possibility that societal changes can drive artistic transformations, as we observed that the context of paintings reflects their societal environment.

\subsection*{Discussion}\label{sec4}
In this study, we investigated how multimodal AI can enhance our understanding of art history by analyzing the contextual elements in paintings. Although these contextual elements are widely recognized as key drivers of artistic evolution~\cite{hauser1951social, gombrich1995story}, they have remained largely unexplored in data-driven studies owing to methodological challenges. Recent advances in AI, particularly through the vision-to-text multimodal approach, have enabled the systematic analysis of painting contexts, as exemplified by the CLIP-based C-vector extraction in this study. Our multimodal AI approach, combined with a massive art dataset~\cite{mao2017deepart}, provided an unprecedented opportunity to quantitatively explore the evolution of human creativity by connecting traditional art historical approaches with data-driven methods.

Our study demonstrated hidden pathways in Western paintings by comparing the formal and contextual aspects, suggesting that artistic evolution is not solely driven by the isolated movement of artists but by collaboration between artists and society. The superior temporal expressibility of C-vectors compared with A-vectors highlights the significant role of transformations in content in shaping the evolution of art (Figs. \ref{fig:fig1} and \ref{fig:fig2}). The distinctive stylistic signatures of individual artists and style periods are reflected more precisely in C-vectors than in A-vectors. We also found remarkable temporal continuity of C-vectors on UMAP, showing linear trajectories across periods, which indicates that artistic contexts evolve progressively.

Additionally, our procedure extends culturomics~\cite{michel2011quantitative} to the visual arts domain by combining C-vectors and generative AI. We observed that the frequency of contextual elements in paintings reflects societal changes, demonstrating that computational analysis of art can quantitatively capture cultural evolution (see Fig.~\ref{fig:fig3}). Our framework opens new avenues for the unexplored goal of culturomics that previous studies have identified as essential~\cite{michel2011quantitative}: incorporating artwork into culturomic analysis. SDM-based image generation with an infusion of future contextual information revealed that the evolution of artistic expression can be systematically modeled by modulating contextual elements without changing formal elements (Fig.~\ref{fig:fig4}). In other words, even when the same formal aspects remain familiar, changes in context can lead to new artistic expressions, ensuring that painting remains an ever-evolving medium.

This study did have some limitations. First, our dataset primarily consisted of well-known Western artworks, which may introduce geographical and temporal biases. We indeed observed an abruptly increasing volume of artworks in recent years (Fig.~S1). Second, our filtering criteria (as described in Materials and Methods) may have resulted in uneven representation across different periods, styles, and artists. Our work also relied on a specific AI model with finite expressibility (\textit{e.g.,} 77 tokens limitation), indicating that future studies could benefit from advances in AI. Finally, some artworks could have been included in the models' training datasets, which may introduce biases towards these paintings. However, this effect should be limited because the SDM and CLIP are general-purpose models trained on a large-scale dataset.

Despite these limitations, our contextual approach provides a promising direction for data-driven art history research by demonstrating its robust validity and performance. Potential extensions are required to enhance the impact of  study; for instance, analyzing the temporal evolution of specific contextual elements (\textit{e.g.,} human figures and landscapes) regarding their formal representations, extending the analysis to non-Western art (\textit{e.g.,} Korean landscape paintings \textit{Sansu-hwa} and Japanese woodblock prints \textit{Ukiyo-e}), and integration of other large-scale cultural datasets (\textit{e.g.,} Google N-gram~\cite{michel2011quantitative}). Such expansions would answer our fundamental question of how human creativity evolves in response to changing cultural contexts through the evolution of humankind.

\subsection*{Materials and Methods}\label{methods}

\subsubsection*{Artwork dataset}
In this study, we used ART500K~\cite{mao2017deepart}, which is a large-scale artwork dataset containing 766,820 images of artworks and their associated metadata. The dataset comprises paintings collected from Google Arts \& Culture~\cite{GoogleArts}, WikiArt~\cite{WikiArt}, Web Gallery of Art~\cite{WebGallery}, and other sources. The dataset provides comprehensive information on each painting, including 41,096 painters, dates from BC to the 2000s, 726 artistic styles, 114 artists' nationalities, and other attributes.

\subsubsection*{Data Preprocessing}

Because this study aimed to analyze the evolution of Western paintings, we preprocessed the dataset using the following steps. First, we removed paintings without image files from the ART500K dataset because our analysis was based on images. We then extracted the painting date from the \texttt{Date} column of the Art500K metadata. When the painting year was not explicitly described, we extracted the year from the title (\texttt{painting\_name} column), if available. Because approximately 25\% of the artworks lack a specific creation year but contain approximate values (\textit{e.g.}, 1846--1848, c1540, or 1420s), we grouped paintings by decade (\textit{e.g.}, 1846--1848 to 1840). For artworks spanning multiple decades, we selected the final year and performed decadal rounding down (\textit{e.g.}, 1539--1542 to 1540). This decadal grouping had limited impact on our analysis because most artistic style periods typically span 10 or more years~\cite{gombrich1995story}. In this study, we utilized these grouped values to analyze painting years, unless otherwise specified. After these preprocessing steps, paintings without year information were excluded from the analysis. We also filtered the dataset using the metadata labels in the following order: \texttt{Style}, \texttt{Field}, \texttt{Genre}, and \texttt{Nationality}. Using this order, we first removed entries containing removed keywords (Table~S3), and then retained only entries containing the selected \texttt{Field} keywords (Table~S4). If the artist's name contained non-English letters, it was normalized using Python's internal \texttt{unicodedata.normalize()} function.

In this study, we used the structure of the Stable Diffusion Model (SDM) 2.0 (\url{https://github.com/Stability-AI/stablediffusion}). We used the checkpoint trained to generate images of $512\times512$ pixels by default (\texttt{512-base-ema.ckpt}), with all adjustable parameters set to their default values. All figures were resized to $512\times512$ pixels following the model specifications. We removed 3.2\% of the remaining dataset where one dimension was at least twice that of the other to prevent information loss during the resizing process. In addition, we removed ${\sim}8\%$ of the remaining paintings with resolutions lower than $410\times410$ (in other words, each dimension was lower than ${\sim}80\%$ of $512$) to maintain sufficient image quality for analysis. Finally, we restricted the data to years 1500 to 1990 because the number of available artworks before 1500 was insufficient for analysis (see Fig.~S1). The final processed dataset contained 72,447 paintings, consisting of 2,354 painters and 128 conventional style periods. Note that the number of paintings in each decade group increased (see Fig.~S1).

\subsubsection*{Encoding paintings with latent vectors}\label{method:extract}

We extracted two types of latent vectors: the A-vector using SDM's autoencoder, which compresses the original image dimensions from approximately one million ($512\times512\times4=1,048,576$) to $16,384$ dimensions, and the C-vector using the CLIP model checkpoints from SDM2.0, with a dimension of $1024$. These complementary representations enabled us to investigate both the formal and contextual elements of artworks. The autoencoder transforms an input image \textit{x} from RGB space into a latent representation $L_a = E(x)$ by training the model to replicate the input image with a bottleneck structure~\cite{kingma2013auto}. By contrast, CLIP maps both images and text into a shared latent space, where semantically similar content is positioned closer together~\cite{radford2021learning}.

For Fig.~\ref{fig:fig2}D, we selected sample paintings for each projected value of each principal component (PC) using the following process. First, we divided the PC axis into eight equally sized segments using the maximum and minimum values as boundaries. At each segmentation point, we identified candidate paintings with projected values within a $\pm0.015$ range along the PC axis. Finally, we selected the painting with the minimum perpendicular distance to the PC axis among these candidates to minimize the fluctuation effect from other PCs.

\subsubsection*{Estimation of embedded vectors' year predictability using XGBoost}

We evaluated the temporal expressibility of two vector representations by regression analysis using XGBoost, a scalable tree-boosting model~\cite{chen2016xgboost}. For each vector type (A- and C-vectors), we trained separate models using 70\% of the original paintings as training data, with the respective latent vectors as input features to predict the painting years. The predictability of the models was measured using the remaining 30\% of paintings. These trained models were also used to estimate the temporal features of generated images, as shown in Fig.~\ref{fig:fig1}.

\subsubsection*{Extraction of human-interpretable latent contexts from the image}

To extract human-understandable contextual information from paintings, as we observed high expressibility of C-vectors, we employed the CLIP Interrogator~\cite{clipInterrogatorGthub}, a widely used tool for prompt engineering for text-to-image models. CLIP Interrogator is designed to identify optimal generative prompts to resemble the input image. This tool integrates the CLIP model with the BLIP2~\cite{li2023blip} image captioning model using a two-step process: first, an initial caption was generated using BLIP2, and then it was iteratively enhanced by incorporating pre-defined descriptive tokens (namely, flavors) to optimize the final prompt. There are five flavor categories: \texttt{base flavors, artist names, media, movements,} and \texttt{negative}. However, we excluded the \texttt{artist names} from the original flavor list because they directly indicate a specific artist's style. If the prompt contained non-English letters, the name was normalized using Python's internal \texttt{unicodedata.normalize()} function. All words composed of characters other than English letters and numbers were excluded from subsequent analysis. We measured the raw frequency using the number of unique words in each painting by counting duplicate words within a prompt (for the same artwork) as a single occurrence (for Figs.~\ref{fig:fig3} and S14 and Tables~S1 and S2.

\subsubsection*{Generative experiment with SDM}

We investigated whether the observed temporal contextual patterns guided the evolutionary trajectory of Western paintings by designing a simple experiment using SDM's image-to-image generation. We hypothesized that synthesizing paintings from a specific period within the primary contexts of the subsequent period may generate paintings in the style of the succeeding period. The experiment was performed using 500 paintings sampled from each century, generating new images from these source paintings using two distinct prompt conditions: 1) representative tokens from the consecutive century and 2) an empty prompt (\texttt{`'}) serving as a null model without contextual guidance. We then predicted the estimated year of the generated images using the pre-trained XGBoost regression model.

To identify the representative tokens for each century, we employed TF-IDF, calculated using \texttt{scikit-learn}'s \texttt{TfidfVectorizer}. Here, the TF-IDF score for text $t$ in document $d$ (where each document is a painting in this study) is as follows:

\begin{equation} 
tf\text{-}idf(t, d)=tf(t, d) \times idf(t),
\end{equation} 

\noindent where $tf(t, d)$ represents the term frequency of text $t$ in document $d$. The inverse document frequency $idf(t)$ is defined as

\begin{equation}
idf(t)=\log\frac{1+n}{1+df(t)} + 1, 
\end{equation} 

\noindent where $n$ represents the total number of distinct paintings in the dataset, and $df(t)$ denotes the number of paintings containing the term $t$. We first calculated the TF-IDF values of words in the generated prompt, where an artwork corresponds to a document. These values were then aggregated by computing the sum of the TF-IDF scores for each word across 100-year periods. We determined the word's representative century based on the period in which it showed the highest TF-IDF value. We then selected the top 100 words using the TF-IDF score for each century. After sorting these words using their TF-IDF values, we manually filtered out words that may directly inject style information: 1) artist names, 2) names of art movements or style periods, and 3) numerical values (which might directly reference specific years). From the remaining words, we selected the top $77$ words because CLIP's text encoder has a maximum token length of 77~\cite{radford2021learning}. Because the minimum number of tokens that can be generated from 77 words is exactly 77, selecting this number guaranteed the utilization of ``at least'' 77 tokens (see Table~S5 for the detailed word list). We then generated space-separated prompts from these tokens as ``\texttt{keyword\textsubscript{1} keyword\textsubscript{2} $\cdots$ keyword\textsubscript{77}}'' for Fig.~\ref{fig:fig4}. We also tested a comma-separated version of the prompt as ``\texttt{keyword\textsubscript{1}, keyword\textsubscript{2}, $\cdots$, keyword\textsubscript{77}}'', with the results shown in Fig.~S16. Because commas are counted as separate tokens, the actual number of keywords used in the comma-separated version was reduced. However, commas may prevent words from being interpreted as connected phrases, providing clearer semantic boundaries between individual keywords. Note that any tokens appearing after this limit are automatically ignored by the model.

The strength parameter, defined between 0 for minimum perturbation and 1 for maximum perturbation in the SDM image-to-image model, determines the degree of transformation from the source image to the generated output. This parameter is used internally to control the noise addition process within the DDIM Solver~\cite{song2021denoising}, which serves as a noising--denoising process. Specifically, it first establishes the total number of steps required to reach the maximum noise level (\textit{i.e.}, DDIM steps) and then controls how many of these processes will be applied to the source image (\textit{i.e.}, the diffusion steps). Here, the diffusion step is determined as the product of DDIM steps and strength; naturally, this value is rounded when not an integer. Consequently, changes in strength do not affect the degree of perturbation if the difference is less than $1/\text{(DDIM steps)}$. Therefore, we adjusted the degree of transformation by directly changing the diffusion step rather than modifying the strength parameter with fixed DDIM steps ($=50$).

We also calculated the temporal axis $\textbf{v}_{1500s \rightarrow 1900s}$ as the difference between the average C-vectors of all paintings from the 1500s ($\textbf{v}_{1500s}$) and that from the 1900s ($\textbf{v}_{1900s}$), \textit{i.e.}, $\textbf{v}_{1500s \rightarrow 1900s} = \textbf{v}_{1900s}-\textbf{v}_{1500s}$. This vector represents the directional shift in contextual features across these periods.

\clearpage

\bibliography{references}
\bibliographystyle{sciencemag}

\section*{Acknowledgments}
We thank Sang Hoon Lee and Jiyi Ryu for their insightful discussions.

\paragraph*{Funding:}
This research was supported by the MSIT (Ministry of Science and ICT), Republic of Korea, under the Innovative Human Resource Development for Local Intellectualization support program (IITP-2025-RS-2022-00156360) supervised by the IITP (Institute for Information \& Communications Technology Planning \& Evaluation). This work was also supported by the National Research Foundation of Korea (NRF), funded by the Korean government (grant Nos. NRF-2022R1C1C2004277 (T.Y.) and 2022R1A2C1091324 (J.Y.)). This research was also supported by the Global Humanities and Social Sciences Convergence Research Program through the National Research Foundation of Korea(NRF), funded by the Ministry of Education (2024S1A5C3A02042671 (J.Y.)). The Korea Institute of Science and Technology Information (KISTI) also supported this research by providing KREONET, a high-speed Internet connection.

\paragraph*{Author contributions:}
All authors conceived and designed the analysis, collected the data, contributed data or analysis tools, and wrote the paper. J.K. performed the analysis.

\paragraph*{Competing interests:}
There are no competing interests to declare.

\paragraph*{Data and materials availability:}
ART500K dump for the main analysis are available from the ART500K dataset download page (\url{https://deepart.hkust.edu.hk/ART500K/art500k.html}). Stable Diffusion Version 2, which is also composed of CLIP and autoencoder used in the analysis, is available from its official GitHub repository \url{https://github.com/Stability-AI/stablediffusion}. All source code used to preprocess, analysis, and create the figures is also available from GitHub (\url{https://github.com/aljinny/art-history}).

\end{document}